%% file: root.tex
\documentclass[letterpaper, 10 pt, conference]{ieeeconf}
\IEEEoverridecommandlockouts

\usepackage{booktabs}
\usepackage{mathtools}
\usepackage{arydshln}   
\usepackage{adjustbox} 

\usepackage{amssymb}
\usepackage{gensymb}
\usepackage{stmaryrd}
\usepackage{cite}
\usepackage{color}
\usepackage{multirow}
\usepackage{graphicx,times}
\usepackage{epstopdf}
\usepackage{indentfirst}
\usepackage{amsmath}
\usepackage{amsfonts}
\usepackage{txfonts}
\usepackage{mathrsfs}
\usepackage{theorem}
\usepackage{float}
\usepackage{subfigure}
\usepackage[font=footnotesize]{caption}

\usepackage{verbatim}
\usepackage{hhline}     		
\usepackage{algorithmic}
\usepackage{algorithm}
\usepackage{array}
\usepackage{url}
\usepackage{bm}
\usepackage{xcolor}
\usepackage{microtype}
\usepackage{cleveref}
\usepackage{balance}
\newcolumntype{G}{@{\hspace{3pt}}c@{\hspace{10pt}}c@{\hspace{10pt}}c@{\hspace{12pt}}}

\crefname{section}{\S}{\S\S}
\Crefname{section}{\S}{\S\S}

\newcommand{\ie}{{\em i.e.}}
\newcommand{\eg}{{\em e.g.}}

\DeclareMathOperator*{\argmax}{arg\,max}

\title{MOGS: Monocular Object-guided Gaussian Splatting in Large Scenes}

\author{Shengkai Zhang$^{1}$, Yuhe Liu$^{1}$, Jianhua He$^{2}$, Xuedou Xiao$^{1}$, Mozi Chen$^{1, *}$, Kezhong Liu$^{1}$ 
\thanks{Authors$^{1}$ are with State Key Laboratory of Maritime Technology and Safety, Wuhan University of Technology, Wuhan, China.
        {\tt\small \{shengkai, yuheliu, xuedouxiao, chenmz, kzliu\}@whut.edu.cn}}
\thanks{Author$^{2}$ is with University of Essex, Colchester, U.K.
		{\tt\small j.he@essex.ac.uk}}%
\thanks{$^{*}$Corresponding author: Mozi Chen (\tt\small chenmz@whut.edu.cn).}
}

\begin{document}

\maketitle
\thispagestyle{empty}
\pagestyle{empty}

\begin{abstract}
Recent advances in 3D Gaussian Splatting (3DGS) deliver striking photorealism, and extending it to large scenes opens new opportunities for semantic reasoning and prediction in applications such as autonomous driving. Today's state-of-the-art systems for large scenes primarily originate from LiDAR-based pipelines that utilize long-range depth sensing. However, they require costly high-channel sensors whose dense point clouds strain memory and computation, limiting scalability, fleet deployment, and optimization speed. We present MOGS, a monocular 3DGS framework that replaces active LiDAR depth with object-anchored, metrized dense depth derived from sparse visual-inertial (VI) structure-from-motion (SfM) cues. Our key idea is to exploit image semantics to hypothesize per-object shape priors, anchor them with sparse but metrically reliable SfM points, and propagate the resulting metric constraints across each object to produce dense depth. To address two key challenges, \ie, insufficient SfM coverage within objects and cross-object geometric inconsistency, MOGS introduces 1) a multi-scale shape consensus module that adaptively merges small segments into coarse objects best supported by SfM and fits them with parametric shape models, and 2) a cross-object depth refinement module that optimizes per-pixel depth under a combinatorial objective combining geometric consistency, prior anchoring, and edge-aware smoothness. Experiments on public datasets show that, with a low-cost VI sensor suite, MOGS reduces training time by up to 30.4\% and memory consumption by 19.8\%, while achieving high-quality rendering competitive with costly LiDAR-based approaches in large scenes. The source code will be publicly available at \url{https://github.com/ClarenceZSK/MOGS/}.

\end{abstract}

\section{Introduction}
\label{sec:intro}
\input{intro}

\section{Related Work}
\label{sec:related}
\input{related}

\section{System Design of MOGS}
\label{sec:design}
\input{design}

\section{Experimental Evaluation}
\label{sec:evaluation}
\input{eval}

\section{Conclusion}
\label{sec:conclusion}
\input{conclusion}

\section*{Acknowledgement}
\label{sec:ack}
This work was supported in part by the National Natural Science Foundation of China (NSFC) under Grant 52571413, Grant 52031009, Grant 52401423, and Grant 62402353; and in part by the EPSRC with RC under Grant EP/Y027787/1 and UKRI under Grant EP/Y028317/1.

\bibliography{root.bib}
\bibliographystyle{IEEEtran}
\balance

\end{document}

%% file: intro.tex

%
%
3D Gaussian Splatting (3DGS) has recently achieved real-time, photorealistic view synthesis with impressive fidelity~\cite{kerbl20233d,duan20244d,Wu_2024_CVPR}. When extended to large scenes, such high-fidelity mapping holds substantial potential for advanced environment sensing and reasoning, enhancing intelligent applications such as autonomous driving. Among existing approaches, LiDAR-initialized 3DGS is particularly effective at scale thanks to metrically accurate ranging and robustness across textures~\cite{hong2024liv,zhao2024tclc,lang2025gaussian,hong2025gs}. However, the high-channel LiDAR typically required is costly and produces dense point clouds; using these clouds to seed Gaussians balloons the number of primitives, inflating memory footprints and slowing training. At the city scale, these hardware and computational burdens constrain fleet deployment and hinder rapid iteration on algorithms and maps. Consequently, despite its accuracy, LiDAR-enabled 3DGS faces practical scalability limits in large scenes.

\begin{figure}[t]
	\centering
	\includegraphics[width=3.4in]{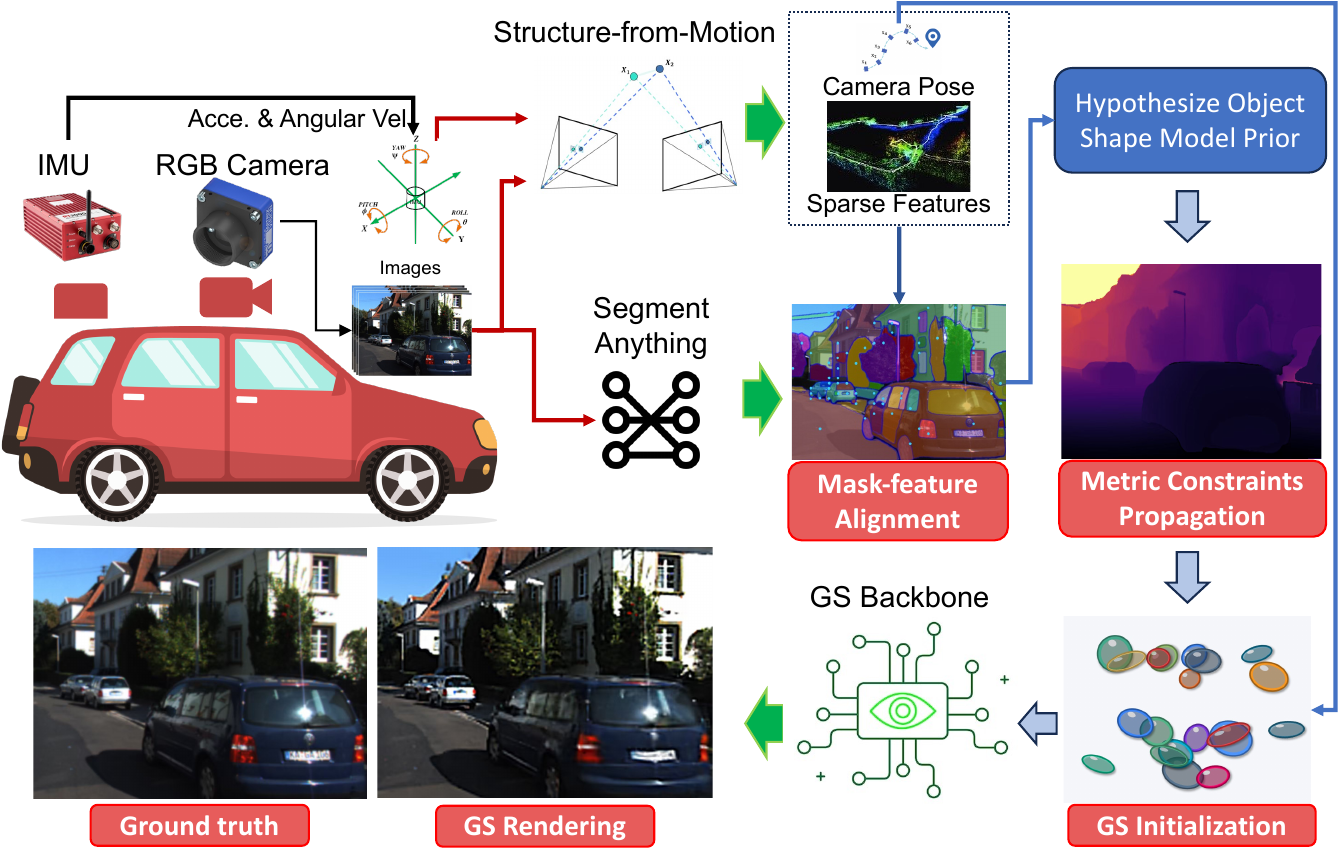}
	\caption{MOGS exploits the rich semantics of RGB images and sparse, metrically reliable SfM cues to infer object-level shape priors. By propagating these metric constraints through shape priors, it produces metrized dense depth, approaching the effect of costly high-channel LiDAR, thereby enabling better Gaussian initialization and higher-quality rendering.} 
	\label{fig:toy}  
	\vspace{-3mm}
\end{figure}

This motivates a monocular pathway to 3DGS in large scenes that is low-cost and unconstrained in range. The primary barrier to monocular 3DGS is the lack of reliable metric depth to initialize Gaussians, leading to scale drift and geometric inconsistency in large scenes~\cite{wu2025vings, matsuki2024gaussian, hu2025splatmap, cheng2025outdoor}. We propose MOGS, a monocular 3DGS framework for large scenes with a visual-inertial (VI) structure-from-motion (SfM) frontend as shown in Fig.~\ref{fig:toy}. At its core is an object-anchored metrization strategy that leverages image semantics to hypothesize shape priors, aligns them with sparse but metrically reliable SfM cues~\cite{vinsmono, cao2022gvins}, and propagates the resulting metric constraints across each object. Thus, we amplify sparse anchors into dense depth support.

However, realizing this idea poses two significant challenges. The first challenge is insufficient SfM coverage within objects. Monocular SfM yields tracks that cluster on corners and high-texture boundaries, leaving the interiors of large, low-texture surfaces, \eg, roads, sky-occluded roofs, glass, sparsely or unevenly anchored, especially at long range. Occlusions and view-dependent appearance further reduce co-visibility, so per-object geometry becomes under-constrained by a handful of edge points, making any interior shape estimate ill-posed. The other challenge is enforcing geometric consistency across objects. Even when individual objects admit plausible local geometry, their relative placement, such as parallelism, coplanarity, contact with the ground plane, and mutual occlusion ordering, remains weakly constrained by monocular cues. Minor per-object scale errors and segmentation boundary jitter accumulate across views, resulting in inter-object depth offsets and broken discontinuities that violate the global structure.

MOGS consists of two designs to address these challenges.

First, we propose a multi-scale shape consensus module. Starting from Segment Anything~\cite{kirillov2023segment} masks, the module iteratively merges small segments that lack SfM support until each region contains sufficient features. Then it fits a compact set of parametric models, \eg, plane, cylinder, and ellipsoid, and chooses the one with the maximal SfM agreement. The selected model propagates metric depth to all pixels in the region, yielding dense, object-consistent priors with confidence from fit quality. Regions without reliable parametric consensus, \eg, foliage and cluttered surfaces, bypass modeling and are refined in the subsequent per-pixel optimization. 

Second, we introduce a cross-object depth refinement module that refines per-object depth with a three-term combinatorial objective to enforce inter-object consistency. The objective comprises 1) a geometric consistency term that encourages agreement between the propagated metrized depth and the dense monocular depth from a large foundation model (LFM), Depth Anything~\cite{yang2024depthv1,yang2024depth} up to a learned scale; 2) an LFM prior anchoring term that softly penalizes deviations between the refined depth and the LFM estimate, serving as a local-shape prior in weakly constrained or nonparametric regions without overriding metric alignment from other terms; and 3) an edge-aware smoothness term that preserves discontinuities while denoising interiors. Optimizing these terms per object implicitly couples neighboring regions through boundary constraints, yielding a metrically coherent depth field for 3DGS in large scenes.

{\bf Contributions}. MOGS makes three contributions:
\begin{itemize}
	\item We propose a multi-scale shape consensus module that establishes object-level shape models that agree with sparse SfM cues and converts them into dense, metrized depth priors for all pixels in each object, enabling reliable Gaussian initialization in large scenes. 
	\item We develop a cross-object depth refinement module that optimizes per-object depth with a three-term combinatorial objective, \ie, geometric consistency, LFM prior anchoring, and edge-aware smoothness, to align neighboring objects and produce a globally coherent depth field.
	\item Extensive experiments on public datasets demonstrate that MOGS reduces training time by up to 30.4\% and memory consumption by 19.8\%, while achieving rendering quality competitive with costly LiDAR-based approaches in large scenes.
\end{itemize}


%% file: related.tex
The rise of 3D Gaussian Splatting (3DGS) has enabled real-time neural rendering by representing scenes as collections of oriented Gaussian ellipsoids that can be efficiently rasterized~\cite{kerbl20233d}. While early 3DGS methods achieved impressive fidelity on bounded scenes~\cite{yu2024mip, lu2024scaffold}, recent work has pushed toward large-scale reconstruction where scalability and robustness are critical~\cite{liu2024citygaussian, lin2024vastgaussian}. While these approaches commonly partition scenes to maintain real-time performance, they typically require precomputed camera poses and rely on depth priors for stable initialization. In large scenes, obtaining such priors is challenging in that SfM can fail under wide baselines, sparse textures, or repetitive patterns~\cite{vinsmono, schoenberger2016vote}.


To address reliability at scale, a growing body of work fuses LiDAR with cameras to inject robust geometry and stabilize poses~\cite{xie2025gs, hong2024liv, zhao2024tclc, lang2025gaussian}. LIV-GaussMap\cite{hong2024liv} initializes surface Gaussians and frame poses from a LiDAR-inertial backbone, then refines the map with image photometric gradients for real-time rendering across large environments. TCLC-GS\cite{zhao2024tclc} constructs a hybrid explicit and implicit representation, supervising splats with dense mesh depth to achieve fast training and high FPS rendering on Waymo and nuScenes. Gaussian-LIC~\cite{lang2025gaussian} integrates LiDAR-IMU-camera SLAM with Gaussian mapping, initializing Gaussians from both LiDAR and triangulated visual points to reach real-time, photorealistic SLAM in unbounded scenes. Collectively, these systems demonstrate that LiDAR-guided 3DGS scales well and renders efficiently in large scenes. However, dense depth supervision over long ranges often demands high-channel LiDAR, which is costly (thousands of USD), and the resulting dense point clouds inflate memory footprints and slow GS optimization, limiting scalability and iteration speed in practice.

To avoid the cost and computational burden of high-channel LiDAR, recent works have explored monocular pathways for large-scale 3DGS. A typical approach is to leverage LFMs for monocular depth estimation, such as Depth Anything~\cite{yang2024depthv1,yang2024depth}, to produce dense depth maps that initialize Gaussians. ICP-3DGS~\cite{zhang2025icp} couples LFM monocular depth with ICP to estimate poses and expand scenes for unbounded environments. Mode-GS~\cite{lee2024mode} parameterizes anchors to calibrate scales of LFM monocular depth to bootstrap rendering. Complementary efforts in mono-GS SLAM~\cite{wu2025vings, deng2025gigaslam, cheng2025outdoor} use monocular metric depth~\cite{piccinelli2024unidepth} during tracking and mapping to keep global structure consistent over long trajectories. DepthSplat~\cite{xu2025depthsplat} underscores the bidirectional benefit of enforcing photometric and geometric consistency when scaling to large scenes. Nonetheless, these methods rely on fine-tuned LFMs tailored to specific datasets, which limits generalization. 
 
Our object-guided approach fuses the strengths of geometrically consistent, up-to-scale LFM depth with metrically anchored sparse SfM cues. We exploit image semantics to infer object-level shape models and propagate sparse depth across each object, computing dense guidance without storing dense point clouds, thereby reducing memory consumption and computational complexity. Additionally, we utilize the geometric consistency of LFM depth during refinement to enhance accuracy and cross-object coherence in the reconstructed scene.


%% file: design.tex
\subsection{Overview}
\label{subsec:overview}

\begin{figure*}[t]
	\centering
	\includegraphics[width=6in]{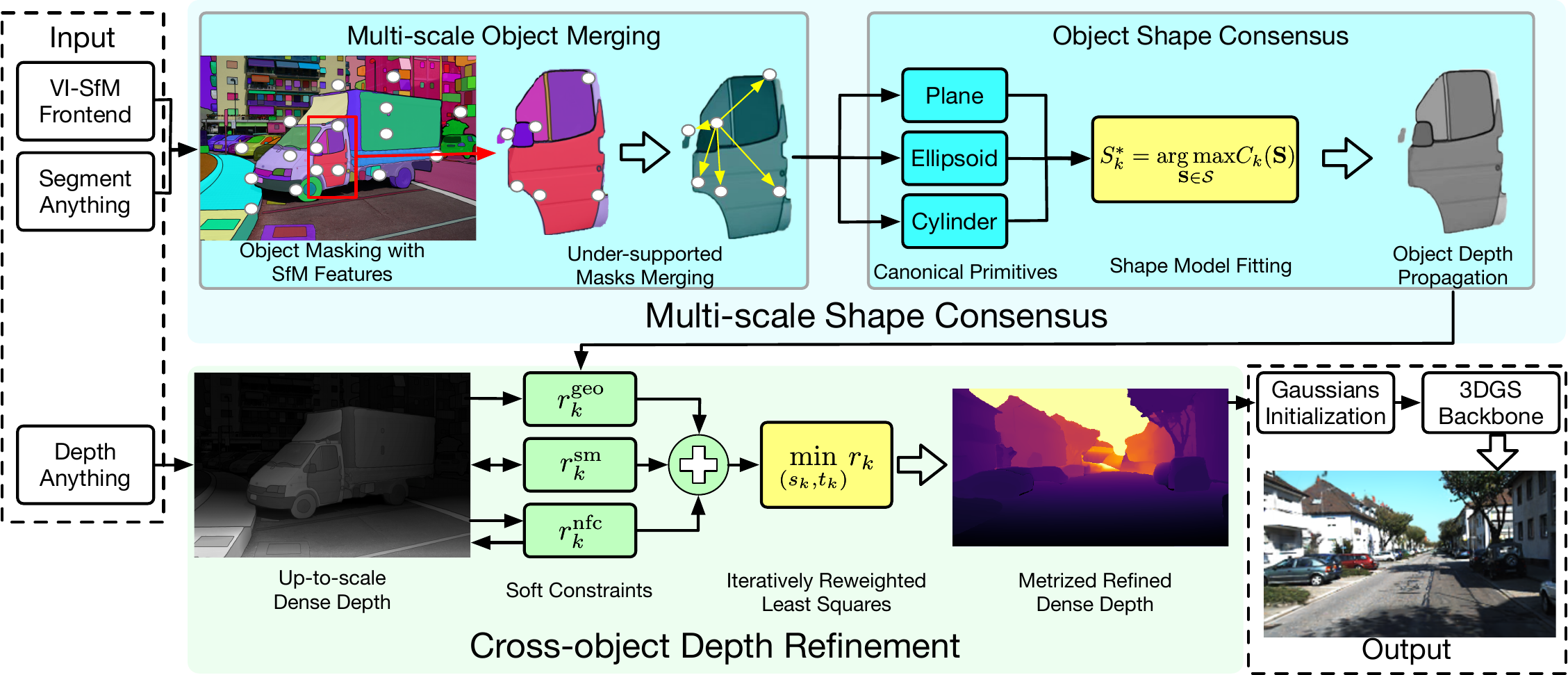}
	\caption{System overview of MOGS. We align VI-SfM visual features with semantic masks from Segment Anything. The multi-scale shape consensus module then establishes object-level shape models and propagates SfM cues to produce metrized dense depth. Building on this, our cross-object depth refinement leverages LFM dense depth to further optimize pixel-wise estimates, yielding strong 3DGS initialization and, ultimately, higher-fidelity Gaussian splatting.} 
	\label{fig:overview}  
	\vspace{-3mm}
\end{figure*}

MOGS enables low-cost and efficient high-definition (HD) 3DGS rendering in large scenes with a VI-SfM frontend, \eg, VINS~\cite{vinsmono}. As shown in Fig.~\ref{fig:overview}, for each input image we first obtain semantic object masks using Segment Anything~\cite{kirillov2023segment}, then associate the 3D positions of SuperPoint features~\cite{detone2018superpoint} from the frontend with the corresponding masks. The multi-scale shape consensus module uses these associations to establish object-level shape models and propagate SfM cues, producing dense metrized depth for all pixels. This depth is initially coarse due to errors remain in nonparametric regions that do not admit standard models and near object boundaries. We therefore introduce a cross-object depth refinement module that further optimizes the depth with a three-term combinatorial objective. It consists of 1) a geometric consistency term that aligns the propagated metrized depth with scale-ambiguous but geometrically consistent dense LFM depth~\cite{yang2024depth}; 2) an LFM prior anchoring term that softly penalizes deviations between the refined depth and the LFM estimate, serving as a local-shape prior in weakly constrained or nonparametric regions without overriding metric alignment from other terms; and 3) an edge-aware smoothness term that preserves discontinuities while denoising interiors. The refined dense depth then initializes Gaussians and bootstraps 3DGS optimization and rendering.



\subsection{Multi-scale Shape Consensus}
\label{subsec:propagation}
Sparse SfM features~\cite{vinsmono} rarely cover the interiors of large, low-texture objects \eg, glass, roads, leaving their geometry under-constrained and depth noisy at long range. Pixel-wise refinement alone cannot reliably inject metricity into these regions, while naive semantic segments are often over-segmented and too small to accumulate sufficient SfM support. This module reframes the problem at the object level, where simple parametric priors, \eg, plane, cylinder, and ellipsoid, can carry metric constraints across many pixels once a minimal set of reliable SfM anchors exists. 

A multi-scale merging strategy is therefore essential. It adaptively fuses adjacent fine segments until each candidate object amasses enough SfM evidence to hypothesize and validate a shape, avoiding both over- and under-segmentation. Establishing such object-level shapes enables robust propagation of metric depth to all pixels within the object, producing dense metrized depth that stabilizes Gaussian initialization, suppresses floaters, and improves long-range fidelity.

We denote the 3D positions of SuperPoint features~\cite{detone2018superpoint} from our SfM frontend~\cite{vinsmono} as $\{\mathbf{P}_i \in \mathbb{R}^3\}_{i=0, 1, \cdots, N}$, with $\mathbf{P}_i=[x_i, y_i, z_i]$, where $N$ is the total number of features and $z_i$ is the depth of feature $i$ in the current camera frame. Meanwhile, for each image, we obtain $K$ fine-grained semantic object masks, $\{M_k\}_{k=1,  \cdots, K}$ using Segment Anything~\cite{kirillov2023segment}. Let $\mathbf{u}_i$ be the 2D projection of $\mathbf{P}_i$ in this image. We assign each feature to the object whose mask contains its projection, \ie, $l_i = k$ if $\mathbf{u}_i \in M_k$, and collect the associated SfM features per object as $\mathcal{F}_k = \{i \; | \; \mathbf{u}_i \in M_k \}$. Features not covered by any mask are treated as background.

{\bf Multi-scale object merging}. For small or low-texture objects, SfM features often concentrate along discontinuous boundaries, making their shapes hard to recover. We therefore impose a minimum-support threshold $n_{min} = \max\{ n_{plane}, n_{ellipsoid}, n_{cylinder} \}$, where $n_{plane}$, $n_{ellipsoid}$, and $n_{cylinder}$ are the minimum numbers of points needed to uniquely determine a plane, an ellipsoid, and a cylinder, which are $3$, $9$, and $7$, respectively. We deem a mask $M_k$ under-supported if $|\mathcal{F}_k| < n_{min}$. 

Under-supported masks seek to merge into adjacent regions to form a coarser object. Let $\mathcal{N}_u$ be the set of masks adjacent to an under-supported mask $M_u$. We seek a neighbor $M_i \in \mathcal{N}_u$ that is depth-contiguous with $M_u$. For each mask $M_i$, we compute the median SfM depth $\bar{z}_i$ over its features. A neighbor $M_i$ is eligible if the depth gap $\Delta z_{i, u} = | \bar{z}_i - \bar{z}_u | \leq \tau_i$, $\tau_i = 0.1\times \bar{z}_i$, \ie, a $10\%$ relative tolerance. This criterion prevents merging adjacent objects at clearly different elevations, \eg, an object on a table versus the floor. Eligible neighbors are then considered as a candidate set $\mathcal{C}_u$ for merging $M_u$. 

Next, among the candidates $\mathcal{C}_u$, we select the mask $M_c$ with the minimum depth gap $\Delta z_{c, u}$ and merge it with $M_u$ to form a new region $M_{u^{\prime}} = M_c \cup M_u$. Then we update the SfM feature set $\mathcal{F}_{u^{\prime}}=\mathcal{F}_{u} \cup \mathcal{F}_{c}$ and test whether $|\mathcal{F}_k| \geq n_{min}$. If not, recompute candidates for $M_{u^{\prime}}$ and continue merging with its remaining adjacent regions. 

{\bf Object shape consensus.} After merging, each object has sufficient SfM features. For object $k$, we then fit the shape that explains the majority of points in $\mathcal{F}_{k}$. Algorithm~\ref{alg:objectid} is the pseudocode of our object shape consensus. 

We consider three canonical primitives, \ie, plane, ellipsoid, and cylinder, parameterized as: 
\begin{itemize}
	\item Plane: $\mathbf{S}_{1}=(\mathbf{n}, d)$ with $\mathbf n\in\mathbb R^3$ a unit normal ($\|\mathbf n\|=1$) and $d\in\mathbb R$ the signed offset;
	\item Ellipsoid: $\mathbf{S}_{2}=(\mathbf{c}, \mathbf{A})$ with $\mathbf c\in\mathbb R^3$ the center and $\mathbf A\in\mathbb R^{3\times3}$ a symmetric positive-definite shape matrix;
	\item Cylinder: $\mathbf{S}_{3}=(\mathbf p,\mathbf u, r)$ with $\mathbf p\in\mathbb R^3$ a point on the axis, $\mathbf u \in\mathbb R^3$ a unit axis direction ($\|\mathbf u\|=1$), and $r > 0$ the radius.
\end{itemize}

We denote the set of canonical primitives as $\mathcal{S} = \{\mathbf{S}_{1}, \mathbf{S}_{2}, \mathbf{S}_{3}\}$. For each $\mathbf{S} \in \mathcal{S}$, we fit the object's SfM points $\mathcal{F}_k$ using minimal-sample RANSAC~\cite{raguram2013usac}, which returns an inlier set and a confidence score $C_k(\mathbf{S})$. The final model is chosen by
\begin{equation}
	S_k^*=\argmax_{\mathbf{S} \in \mathcal{S}} C_k(\mathbf{S}).
\end{equation}

Specifically, for each $\mathbf{S} \in \mathcal{S}$, in the RANSAC, we repeat sampling up to \(\emph{maxIter}\) and keep the hypothesis that maximizes the number of inliers for the object. We define the inlier by computing the distance of a point $\mathbf{P} \in \mathcal{F}_k$ to a model $\mathbf{S}$ as
\begin{equation}
	r_k(\mathbf{P} \; | \; \mathbf{S}) = \left| \frac{f_k(\mathbf{P}, \mathbf{S})}{\big\| \nabla f_k(\mathbf{P}, \mathbf{S}) \big\|^2 }\nabla f_k(\mathbf{P}, \mathbf{S})  \right|,
\end{equation}
where $f_k(\mathbf{P}, \mathbf{S})$ is the implicit function of the shape model of object $k$ parameterized by $\mathbf{S}$. Points $\mathbf{P}_{0}$ on the surface satisfy $f_k(\mathbf{P}_{0}, \mathbf{S}) = 0$. $\nabla f_k(\mathbf{P}, \mathbf{S})$ denotes its spatial gradient. Given a geometric error threshold $T_s$, the inlier set of object $k$ is
\begin{equation}
	\mathcal{I}_k (\mathbf{S}) = \big\{\mathbf{P} \in \mathcal{F}_k\; \big|\; r_k(\mathbf{P} \; | \; \mathbf{S}) < T_s\big\}.
\end{equation}
Then the inlier ratio can be defined as $\gamma_k(\mathbf{S}) = \frac{| \mathcal{I}_k (\mathbf{S}) |}{| \mathcal{F}_k | }$.

To compute the confidence score $C_k(\mathbf{S})$, we still need to know the residual variance over inliers as
\begin{equation}
	\sigma^2_k (\mathbf{S}) = \frac{1}{| \mathcal{I}_k (\mathbf{S}) |}\sum_{\mathbf{P} \in \mathcal{I}_k (\mathbf{S})} r_k(\mathbf{P} \; | \; \mathbf{S})^2.
\end{equation}
We then define the confidence score, which rewards a high inlier ratio and a low residual variance, as
\begin{equation}
	C_k(\mathbf{S}) = \gamma_k(\mathbf{S}) \exp \!\left(-\,\frac{\sigma_k^2(\mathbf{S})}{\beta\,T_s^2}\right),
	\label{eqn:score}
\end{equation}
where $\beta\in(0,1]$ is a scale hyperparameter. 

At this stage, depths for pixels on structured objects can be readily propagated from their fitted shape models. In contrast, points not explained by any canonical primitive typically lie in texture-rich regions, \eg, foliage or cluttered surfaces. These pixels are sufficiently dense to seed Gaussians, and their depths are further refined in the subsequent module. In practice, the three canonical models cover most pixels in large scenes: on street-scene data, roughly 44\% of segments (covering 73\% of pixels) are assigned to our geometric structures.

\begin{algorithm}[t]
\caption{Object Shape Consensus}
\label{alg:objectid}
\begin{algorithmic}[1]
  \STATE Input: the SfM points $\mathcal{F}_k$ on object $k$, $k=1, 2, \cdots, K$.
  \STATE $S \gets \emptyset$
  \FOR{$k = 1$ \TO $K$}
    \STATE $S_k^* \gets \emptyset$, \quad$\mathcal{I}_k^{\text{best}} \gets \emptyset$, \quad $C_k^{\text{best}} \gets -\infty$,
    \FOR{each $\mathbf{S} \in \mathcal{S}$}
      \STATE $R_c \gets $ \textsc{MinimalSampleSize}($\mathbf{S}$)
      \STATE $C_k(\mathbf{S}) \gets -\infty$,\quad $\mathcal{I}_k^{\text{best}} (\mathbf{S}) \gets \emptyset$
      \FOR{$j = 1$ \TO $\emph{maxIter}$}
        \STATE $r_k \gets $ \textsc{RandomSample}($\mathcal{F}_k$, $R_c$)
        \STATE $S_k \gets $ \textsc{FitStructure}($\mathbf{S}$, $r_k$)
        \STATE $\mathcal{I}_k (\mathbf{S}) \gets \{\mathbf{P} \in \mathcal{F}_k \;|\; r_k(\mathbf{P} \; | \; \mathbf{S}) < T_s\}$
        \STATE $score \gets Eqn.~\eqref{eqn:score}$
        \IF{$score > C_k(\mathbf{S})$}
          \STATE $C_k(\mathbf{S}) \gets score$,\quad $S_k^* \gets S_k$
          \STATE $\mathcal{I}_k^{\text{best}} (\mathbf{S}) \gets \mathcal{I}_k (\mathbf{S})$
        \ENDIF
      \ENDFOR
      \IF{$C_k(\mathbf{S})  > C_k^{\text{best}}$}
        \STATE $S_k^* \gets S_k$,\quad $ \mathcal{I}_k^{\text{best}} \gets \mathcal{I}_k^{\text{best}} (\mathbf{S})$, \quad $C_k^{\text{best}} \gets C_k(\mathbf{S})$
      \ENDIF
    \ENDFOR
    \IF{$|\mathcal{I}_k^{\text{best}}| \geq \tau \cdot |\mathcal{F}_k|$}
      \STATE $S \gets S \cup \{S_k^*, C_k^{\text{best}})\}$
    \ENDIF
  \ENDFOR
  \RETURN $S$
\end{algorithmic}
\end{algorithm}

\subsection{Cross-object Depth Refinement}
The depth propagated in the previous module can inherit SfM errors and shape-model bias, and it lacks cross-object constraints to preserve global structure. To address this, we incorporate the LFM, Depth Anything~\cite{yang2024depth}, which provides scale-ambiguous yet geometrically consistent dense depth from monocular images. We use this LFM depth as cross-object geometric supervision to refine the propagated depths.

Given a mask \(M_k\) with pixel domain \(\Omega_k\), we map the LFM's relative depth \(\hat d(p)\) at pixel \(p=(u,v)\) to absolute depth by an affine calibration:
\begin{equation}
\tilde D_k(p) \;=\; s_k\,\hat d(p) + t_k,\qquad p\in\Omega_k,
\label{eq:affine-depth}
\end{equation}
where \((s_k,t_k)\) are mask-wise scale and shift.

Let $D_k^{\text{model}}(p)$ be the depth at pixel $p$ obtained by propagating the fitted canonical model for mask $M_k$, and let $C_k \in [0,1]$ denote that model's confidence. We estimate the affine calibration $(s_k, t_k)$ with three complementary terms evaluated over the same domain $\Omega_k$. A robust Huber loss $\rho(\cdot)$ downweights outliers.

{\em Geometric consistency residual}. This term enforces agreement between the refined \(\tilde D_k\) and the model-propagated depth \(D_k^{\text{model}}\):
\begin{equation}
\label{geo-consis}
r_{k}^{\mathrm{geo}} = \sum_{p\in\Omega_k} C_k\times \rho \big(\tilde D_k(p)-D_k^{\text{model}}(p)\big).
\end{equation}

{\em LFM prior anchoring residual}. This term softly anchors \(\tilde D_k(p)\) to the LFM estimate \(\hat d(p)\), helping preserve local structure in regions with weak geometric support:
\begin{equation}\label{aff-consis}
r_{k}^{\mathrm{nfc}} = \sum_{p\in\Omega_k} \rho \big(\tilde D_k(p)-\hat d(p)\big).
\end{equation}

{\em Edge-aware smoothness residual}. This term encourages intra-mask smoothness while respecting image gradients to preserve sharp object boundaries.
\begin{equation}\label{edge-smoo}
r_{k}^{\mathrm{sm}} = \sum_{(p,q)\in\mathcal{N}_k} w_{pq}\,\bigl(\tilde D_k(p)-\tilde D_k(q) \bigr)^2.
\end{equation}
where $\mathcal{N}_k$ is a pixel neighborhood over $\Omega_k$ and $w_{pq}$ is edge-gated pairwise weight that depends on local image gradients. 
\begin{equation}\label{eq:w-edge}
    w_{pq} = \exp \Big(-\tfrac{\|\nabla I(p)\|_2^2+\|\nabla I(q)\|_2^2}{2\,\sigma_I^2} \Big), (p,q) \in \mathcal{N}_k
\end{equation}
where \(\nabla I(\cdot)\) denotes the image spatial gradient, and \(\sigma_I\) controls the aggressiveness of edge downweighting. $(p,q) \in \mathcal{N}_k$ restricts smoothing within mask $M_k$, preventing cross-mask diffusion. The weight \(w_{pq}\in(0,1]\) decays near strong gradients, yielding anisotropic smoothing.

For mask $M_k$, we optimize the following combinatorial objective:
\begin{equation}\label{Joint-loss}
\min_{(s_k,t_k)}\ 
r_{k} = \lambda_{\text{geo}} r_{k}^{\mathrm{geo}} + \lambda_{\text{nfc}} r_{k}^{\mathrm{nfc}} + \lambda_{\text{sm}} r_{k}^{\mathrm{sm}},
\end{equation}
where the positive weights $\lambda_{\text{geo}}$, $\lambda_{\text{nfc}}$, and $\lambda_{\text{sm}} > 0$ balance the geometric consistency, LFM prior anchoring, and edge-aware smoothness terms, respectively. We set them empirically per scene. In outdoor driving experiments, we use $\lambda_{\text{geo}} = 1.0$, $\lambda_{\text{nfc}} = 0.7$, and $\lambda_{\text{sm}} = 0.1$. For scenes dominated by large planes and regular structures, $\lambda_{\text{geo}}$ can be increased, \eg, to $1.2$. For low-texture, repetitive-texture, or cluttered scenes, we recommend raising $\lambda_{\text{nfc}}$, \eg, to $0.9$. For foggy or low-light conditions with noisy depth, increasing $\lambda_{\text{sm}}$, \eg, to $0.2$, helps suppress artifacts. 

We solve \eqref{Joint-loss} by iteratively reweighted least squares (IRLS) and compute the refined depth via \eqref{eq:affine-depth}.

%% file: eval.tex
\subsection{Implementation and Experimental Setup}
MOGS aims to produce efficient, reliable dense depth maps that enable stronger Gaussian initialization and high-fidelity 3DGS rendering in large scenes. Accordingly, we evaluate MOGS on depth accuracy and final rendering quality.

{\bf Implementation.}
All experiments run on a single NVIDIA GeForce RTX~4090 GPU and an Intel Xeon Platinum~8347C CPU (2.10 GHz). We detect SuperPoint features and match them with LightGlue; correspondences are filtered by USAC-PROSAC. Accepted keypoints carry persistent global track IDs for downstream reconstruction. Keyframes are triggered jointly by inlier count and parallax thresholds. We initialize 3D points via SVD triangulation and refine them with nonlinear least squares using analytic Jacobians. 

Depth-aware instance masks are sorted far-to-near by median depth, refined with a 2-pixel neighborhood voting radius, small-component merging to adjacent structures. Given an LFM depth map $r$, each mask $M_k$ is calibrated by an affine mapping $\tilde D_k = s_k r + t_k$. Masks with $\text{mean}(r) > \tau_{\text{sky}} = 100$ are treated as sky and clamped to a maximum depth. When sufficient SfM points are present, $\{s_k,t_k\}$ are estimated via IRLS. Note that dynamic features in a scene are identified and compensated by~\cite{fan2024enhancing}.



{\bf Datasets and Metrics.}
We use KITTI-Depth~\cite{geiger2013vision} collection that covers urban, residential, road, and campus traffic scenes. The raw sensor platform comprises two global-shutter stereo rigs, a Velodyne HDL-64E LiDAR, and an OXTS RT3003 high-precision IMU; cameras and LiDAR are synchronized at 10 Hz, and the IMU at 100 Hz. KITTI-Depth provides approximately 93\,k semi-dense depth maps temporally aligned with RGB camera and LiDAR. In addition, we also use KITTI-360~\cite{liao2022kitti} that contains large-scale suburban cruising (total distance 73.7 km, over 83 k frames). The platform carries a forward-looking perspective stereo pair plus fisheye cameras, a Velodyne LiDAR, and SICK push-broom 2D lasers, providing 360$^{\circ}$ perception with consistent semantic labels. Ground-truth camera poses are retrieved from the datasets or captured via a motion capture system like NOKOV using customized experimental platforms. Correspondingly, all ground-truth depth maps were temporally and spatially aligned with the captured images.


Depth accuracy is reported using Absolute Relative Error ($AbsRel$), Root Mean Squared Error ($RMSE$), and $\delta_1$ (percentage of $\max(d^*/d, d/d^*) < 1.25$). For 3DGS, rendering quality is measured by Peak Signal-to-Noise Ratio ($PSNR$), Structural Similarity Index ($SSIM$), and Learned Perceptual Image Patch Similarity ($LPIPS$). Training efficiency is measured by optimization iterations and wall-clock time to reach a target quality level under a fixed budget, and computational cost is further characterized by the number of active Gaussian primitives and peak GPU memory footprint.

\subsection{Performance Evaluation}
\label{sec:evalu}
\textbf{Depth Accuracy.}
We compare against SOTA monocular-based depth estimation baselines \emph{Depth Anything V2}~\cite{depth_anything_v2}, \emph{Depth Pro}~\cite{bochkovskii2024depth}, \emph{ZoeDepth}~\cite{bhat2023zoedepth}, and \emph{Metric3D v2}~\cite{hu2024metric3d}. All these methods infer dense depth from monocular images using a pre-trained LFM. In the evaluation, we use the ground truth in the dataset as a reference to compute the average metrics.
Quantitative results are summarized in Table~\ref{tab:depth_main}, showing that MOGS attains consistently lower $AbsRel$ and $RMSE$, and higher $\delta_1$. Notably, the metrized dense depth obtained by MOGS surpasses all LFM estimates without any pre-training.



\begin{table}[t]
\centering
\caption{Monocular depth estimation on KITTI-Depth and KITTI-360.}
\label{tab:depth_main}
\footnotesize
\setlength{\tabcolsep}{2pt} 
\renewcommand{\arraystretch}{1.20}
\begin{tabular}{lcccccc}
\toprule
\multirow{2}{*}{Method} &
\multicolumn{3}{c}{KITTI-Depth} &
\multicolumn{3}{c}{KITTI-360} \\
\cmidrule(lr){2-4}\cmidrule(lr){5-7}
& AbsRel$\downarrow$ & RMSE$\downarrow$ & $\delta_1\uparrow$
& AbsRel$\downarrow$ & RMSE$\downarrow$ & $\delta_1\uparrow$ \\
\midrule
Depth Anything V2 & 0.070 & 2.62 & 0.958 & 0.080 & 2.99 & 0.948 \\
Depth Pro          & 0.060 & 2.34 & 0.970 & 0.071 & 2.70 & 0.959 \\
ZoeDepth           & 0.082 & 3.10 & 0.941 & 0.093 & 3.55 & 0.929 \\
Metric3D v2        & 0.059 & 2.32 & 0.971 & 0.070 & 2.67 & 0.960 \\
\textbf{MOGS (Ours)} &
\textbf{0.058} & \textbf{2.29} & \textbf{0.973} &
\textbf{0.070} & \textbf{2.63} & \textbf{0.961} \\
\bottomrule
\end{tabular}
\end{table}

\begin{table}[t]
\centering
\caption{NVS quality and convergence under fixed 3DGS budgets.}
\label{tab:nvs_convergence}
\footnotesize
\setlength{\tabcolsep}{4pt}
\renewcommand{\arraystretch}{1.20}
\begin{tabular}{lccccc}
\toprule
Init & Iter.$\downarrow$ & Gauss. ($\times10^6$) & PSNR$\uparrow$ & SSIM$\uparrow$ & LPIPS$\downarrow$ \\
\midrule
Init-Rand      & 46k & 1.67 & 19.7 & 0.936 & 0.332 \\
Init-LFM-M    & 39k & 1.53 & 21.1 & 0.951 & 0.288 \\
\textbf{Init-MOGS} & \textbf{32k} & \textbf{1.34} & \textbf{21.9} & \textbf{0.958} & \textbf{0.236} \\
\bottomrule
\end{tabular}
\end{table}

{\bf Rendering Quality.}
We first measure how the depth-enabled initializations affect 3DGS convergence speed, memory consumption, and rendering quality. We then compare our method against state-of-the-art monocular 3DGS baselines for qualitative rendering in outdoor large scenes.

\begin{figure}[t]
	\centering
	\includegraphics[width=\linewidth]{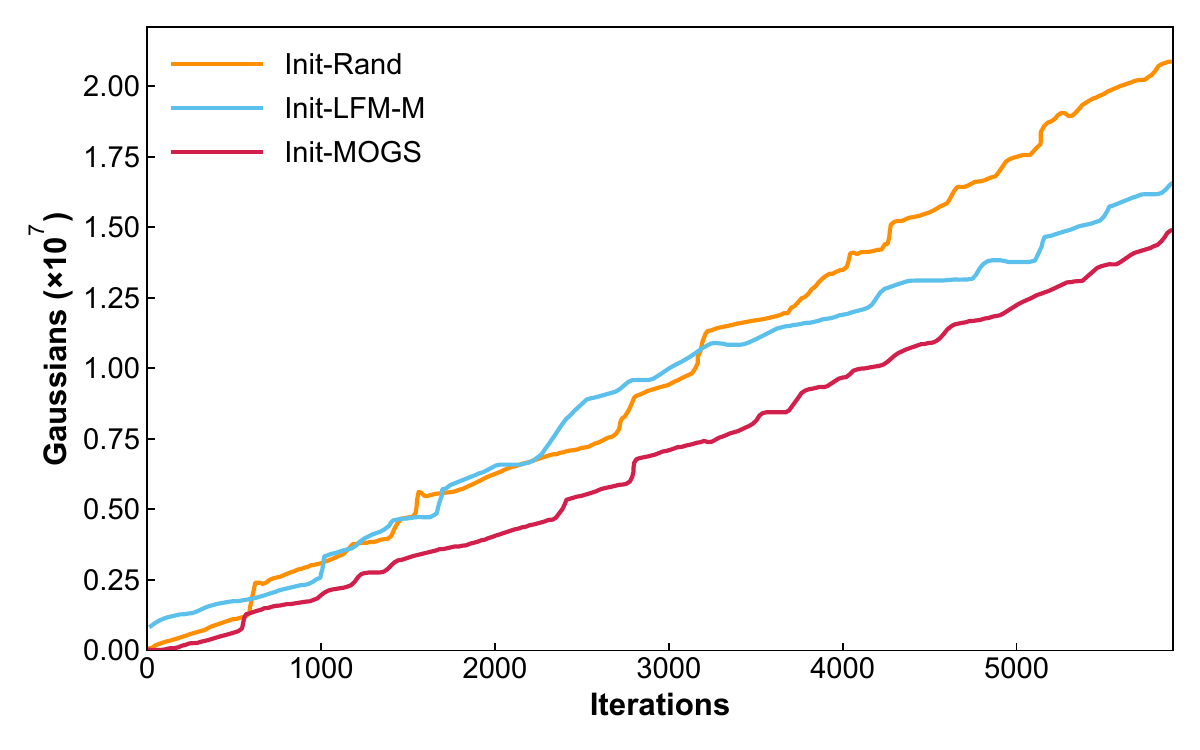}
    \caption{Convergence and efficiency under different GS initializations, revealing how each method grows its Gaussians along with iterations.}
	\label{fig:grow}  
\end{figure}

\begin{figure}[t]
	\centering
	\includegraphics[width=\linewidth]{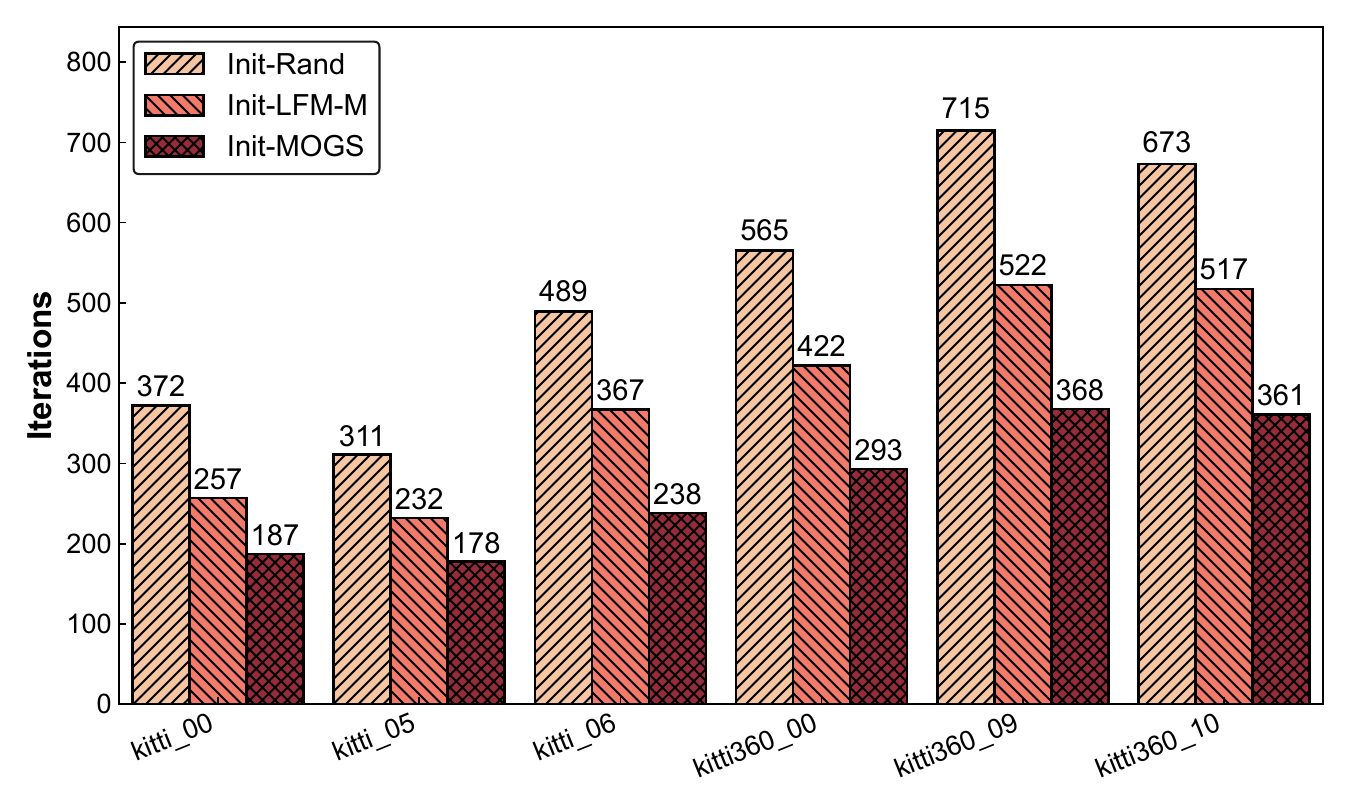}
    \caption{The numbers of iterations under different GS initializations that optimize a view rendering until 20 dB $PSNR$ on multiple datasets.}
	\label{fig:qvsteps}  
\end{figure}

\emph{1) GS initialization.}
We adopt an improved version of MonoGS~\cite{matsuki2024gaussian} as GS backend. We compare different depth-enabled initializations to bootstrap GS optimization and rendering, including \emph{Init-Rand} that initializes with random depth, \emph{Init-LFM-M} that provides a scale factor computed by aligning LFM to ground truth and then metrize the LFM depth, and \emph{Init-MOGS}. 
We conduct this experiment in two rounds. First, we fix a target $PSNR$ and run these methods to see the number of iterations and generated Gaussians (column 2-3 in Table~\ref{tab:nvs_convergence}). Init-MOGS reduces up to $30.4\%$ of iterations and $19.8\%$ of Gaussians to achieve high-quality GS rendering. Second, we fix a target number of iterations and run to see the rendering quality (column 4-6 in Table~\ref{tab:nvs_convergence}). Init-MOGS shows the best rendering quality in terms of $PSNR$, $SSIM$, and $LPIPS$.

In addition to the average metrics in Table~\ref{tab:nvs_convergence}, we plot the Gaussians-iteration curve, illustrating how each method grows its Gaussians during training as shown in Fig.~\ref{fig:grow}. MOGS exhibits the lowest slope, indicating the slowest growing speed. 

Fig.~\ref{fig:qvsteps} shows the number of iterations to the target PSNR (TT-PSNR) with respect to different datasets. We fixed a target view for each experiment, and we start counting iterations from its first visibility of the view until its rendering $PSNR$ exceeds $20$ dB. Across all sequences, MOGS attains the target rendering quality with the fewest iterations and the fewest active Gaussians (refer to Fig.~\ref{fig:grow}), indicating faster anchoring of geometry at the correct scale, reduced over-parameterization and drift, and a better quality-efficiency trade-off. We attribute these gains to object-level metric constraints, which make the geometry converge quickly and shift optimization to appearance earlier.

\emph{2) GS rendering.}

\begin{figure*}[t]
	\centering
	\includegraphics[width=\textwidth]{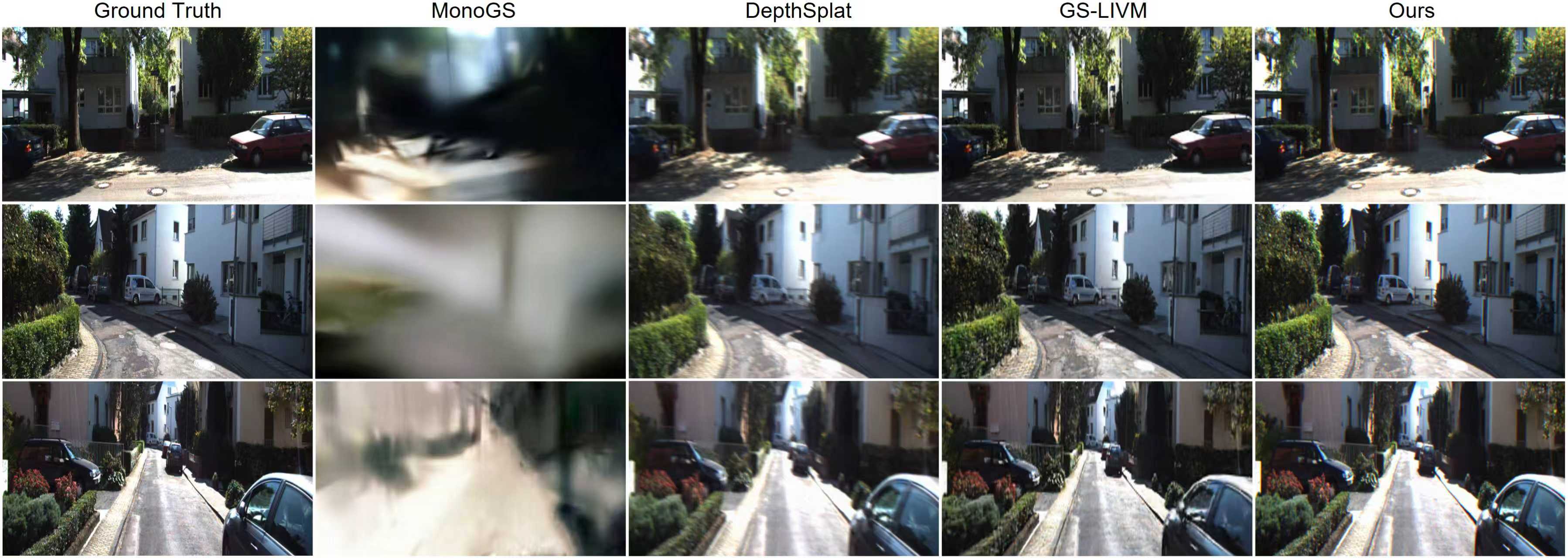}
	\caption{Rendering on KITTI road scenes. Each row shows one representative frame. Columns (left to right) are Ground truth, MonoGS, DepthSplat, GS-LIVM, and MOGS. Our method reconstructs sharper thin structures, preserves depth discontinuities at boundaries, and suppresses long range floaters.} 
	\label{fig:render}  
	\vspace{-3mm}
\end{figure*}

Unless stated otherwise, all runs share identical hyperparameters (training steps, learning rate, cap on visible Gaussians), data splits (held-out novel views), and image resolution. We compare MOGS with SOTA monocular-based GS approaches MonoGS~\cite{matsuki2024gaussian} and DepthSplat~\cite{xu2025depthsplat}, as well as a SOTA LiDAR-based GS approach, GS-LIVM~\cite{xie2025gs}, on our splits using their recommended settings. When the depth is required, we use the method's native inputs. We report rendering metrics under the same computational resource budget. Table~\ref{tab:train_novel} summarizes quantitative novel-view metrics across sequences and Fig.~\ref{fig:render} shows qualitative rendering results on KITTI road scenes. These results indicate that MOGS significantly outperforms monocular-based GS renderers with fewer iterations and lower training cost. Meanwhile, MOGS achieves comparable rendering quality to the SOTA costly LiDAR-enabled baseline, demonstrating the practicality of MOGS. 

\begin{table*}[t]
  \centering
  \caption{Rendering results on outdoor road sequences KITTI.}
  \setlength{\tabcolsep}{8pt}
  \renewcommand{\arraystretch}{1.05}
  \label{tab:train_novel}
  \begin{tabular}{llcccccccccc}
    \toprule
    \textbf{Method} & \textbf{Metric} & \textbf{00} & \textbf{01} & \textbf{02} & \textbf{03} & \textbf{04} & \textbf{05} & \textbf{06} & \textbf{07} & \textbf{08} & \textbf{Avg.} \\
    \midrule
    \multirow{3}{*}{MonoGS}
      & PSNR$\uparrow$    & 14.5 & 12.0 & 14.3 & 16.1 & 16.3 & 15.0 & 14.2 & 16.0 & 13.5 & 14.7 \\
      & SSIM$\uparrow$    & 0.492 & 0.421 & 0.478 & 0.543 & 0.548 & 0.503 & 0.472 & 0.541 & 0.463 & 0.495 \\
      & LPIPS$\downarrow$ & 0.756 & 0.826 & 0.758 & 0.714 & 0.705 & 0.742 & 0.771 & 0.712 & 0.782 & 0.751 \\
    \noalign{\vskip 3pt}\cdashline{1-12}\noalign{\vskip 3pt}
    \multirow{3}{*}{DepthSplat}
      & PSNR$\uparrow$    & 18.1 & 16.5 & 18.1 & 19.9 & 20.1 & 18.7 & 18.2 & 19.7 & 17.8 & 18.6 \\
      & SSIM$\uparrow$    & 0.581 & 0.509 & 0.571 & 0.621 & 0.629 & 0.589 & 0.571 & 0.619 & 0.560 & 0.584 \\
      & LPIPS$\downarrow$ & 0.503 & 0.582 & 0.512 & 0.443 & 0.432 & 0.478 & 0.501 & 0.442 & 0.523 & 0.490 \\
    \noalign{\vskip 3pt}\cdashline{1-12}\noalign{\vskip 3pt}
    \multirow{3}{*}{GS-LIVM}
      & PSNR$\uparrow$    & \textbf{20.4} & \textbf{18.5} & \textbf{20.2} & 21.3 & \textbf{22.1} & 20.4 & 19.4 & \textbf{21.6} & \textbf{19.5} & \textbf{20.4} \\
      & SSIM$\uparrow$    & 0.640 & 0.583 & 0.647 & \textbf{0.706} & 0.689 & \textbf{0.675} & \textbf{0.645} & 0.689 & 0.590 & 0.652 \\
      & LPIPS$\downarrow$ & 0.389 & \textbf{0.502} & 0.381 & 0.309 & 0.312 & 0.363 & \textbf{0.381} & 0.342 & 0.443 & 0.381 \\
    \noalign{\vskip 3pt}\cdashline{1-12}\noalign{\vskip 3pt}
    \multirow{3}{*}{\textbf{Ours}}
      & PSNR$\uparrow$    & 20.1 & 18.1 & 20.1 & \textbf{21.5} & 21.6 & \textbf{20.6} & \textbf{19.6} & 21.3 & 19.0 & 20.0\\
      & SSIM$\uparrow$    & \textbf{0.662} & \textbf{0.603} & \textbf{0.651} & 0.703 & \textbf{0.701} & 0.671 & 0.642 & \textbf{0.702} & \textbf{0.612} & \textbf{0.659} \\
      & LPIPS$\downarrow$ & \textbf{0.387} & 0.503 & \textbf{0.379} & \textbf{0.308} & \textbf{0.309} & \textbf{0.361} & \textbf{0.381} & \textbf{0.339} & \textbf{0.441} & \textbf{0.378} \\
    \bottomrule
  \end{tabular}
\end{table*}

\subsection{Ablation Studies}
We now evaluate the effectiveness of each individual module of MOGS.

{\bf Multi-scale Shape Consensus.}
Removing the multi-scale shape consensus module degrades object coherence on boundaries and small parts, leading to $1.39$ dB $PSNR$ drop and $0.012$ $AbsRel$ on average. The quality-steps curves shift right, and we observe more floaters along depth discontinuities, indicating less reliable Gaussian initialization in weak-texture regions.

{\bf Cross-object Depth Refinement.}
Disabling the depth refinement yields the largest performance degradation ($-0.78$ dB $PSNR$, $-0.010$ SSIM, $0.012$ $AbsRel$), confirming that per-object metrization is critical for anchoring LFM depth and preserving intra-object shape. Removing this refinement module breaks outlier robustness and introduces a per-mask scale offset that propagates to the Gaussian size and placement. The optimizer then overcompensates with more primitives and steps, yet still converges more slowly and with more artifacts across the scene.

As reported in Table~\ref{tab:abalation}, the full MOGS configuration delivers the best quality. Disabling MSC introduces visible floaters near object boundaries and yields a 6.95\% decrease in $PSNR$ and a 1.97\% decrease in $SSIM$, together with a 20.69\% increase in $AbsRel$. This suggests that without multi-scale merging and parametric shape fitting, object-level geometric anchors become unreliable, which leads to poor initialization and unstable optimization. When both MSC and CDR are removed, the effect is larger. $PSNR$ decreases by 15.90\% and $SSIM$ by 3.64\%, while $AbsRel$ increases by 34.48\%, underscoring the importance of cross object depth refinement for global consistency and rendering quality.


\begin{table}[t]
  \centering
  \caption{Ablation results on sequence. Higher is better for PSNR/SSIM, lower is better for AbsRel.}
  \label{tab:ablation}
  \setlength{\tabcolsep}{10pt} 
  \renewcommand{\arraystretch}{1.40}
  \begin{tabular}{lccc}
    \toprule
    Variant & PSNR$\uparrow$ & SSIM$\uparrow$ & AbsRel$\downarrow$ \\
    \midrule
    full MOGS   & \textbf{20.0} & \textbf{0.659} & \textbf{0.058} \\
    w/o MSC      & 18.61 & 0.646 & 0.070 \\
    w/o MSC + CDR     & 16.82 & 0.635 & 0.078 \\
    \bottomrule
  \end{tabular}
  \begin{tabular}{l}
    MSC stands for Multi-scale Shape Consensus; \\  
    CDR stands for Cross-object Depth Refinement.
    \raggedright
  \end{tabular}
  \label{tab:abalation}
\end{table}

%% file: conclusion.tex

This paper presents MOGS, a monocular object-guided pipeline for 3D Gaussian Splatting in large scenes that replaces costly active LiDAR depth with object-anchored, metrized dense depth derived from sparse SfM cues, enabling strong Gaussian initialization and high-fidelity rendering. The core insight is to treat objects as carriers of metric geometry. MOGS has two key components: 1) a multi-scale shape consensus module that extracts object-level primitives from image semantics and propagates SfM depth within each object via these primitives; and 2) a cross-object depth refinement module that yields a metrically coherent depth field through a three-term objective, including geometric consistency with LFM depth, LFM prior anchoring, and edge-aware smoothness. Extensive experiments on public datasets show that, using a low-cost VI sensor suite, MOGS reduces training time and memory footprint while achieving rendering quality competitive with costly LiDAR-based approaches.